\documentclass[letterpaper, 10 pt, conference]{ieeeconf}  

\IEEEoverridecommandlockouts                              

\usepackage{amssymb}    

\usepackage{algorithm}
\usepackage{algorithmic}

\usepackage{graphicx}
\graphicspath{{figures/}}
\usepackage{subfigure}

\usepackage{booktabs}   

\usepackage{mathtools}
\usepackage{array}

\usepackage[switch]{lineno}     

\usepackage{color} 

\usepackage{dutchcal} 

\title{\LARGE \bf
Whole-Body Control with Motion/Force Transmissibility for Parallel-Legged Robot
}

\author{Jiajun Wang$^{1}$, Gang Han$^{1}$, Xiaozhu Ju$^{1}$ and Mingguo Zhao$^{2}$
\thanks{$^{1}$Jiajun Wang, Gang Han, and Xiaozhu Ju are with Beijing Research Institute of UBTECH Robotics, Beijing, China.
        {\tt\small \{jiajun.wang, gang.han, xiaozhu.ju\}@ubtrobot.com}}%
\thanks{$^{2}$Mingguo Zhao is with Department of Automation, Tsinghua University and Beijing Innovation Center for Future Chips, Tsinghua University, Beijing, China.
        {\tt\small mgzhao@mail.tsinghua.edu.cn}}%
}

\begin{document}


\maketitle
\thispagestyle{empty}
\pagestyle{empty}

\begin{abstract}

For achieving kinematically suitable configurations and highly dynamic task execution, an efficient way is to consider robot performance indices in the whole-body control (WBC) of robots.
However, current WBC methods have not considered the intrinsic features of parallel robots, especially motion/force transmissibility (MFT). 
This paper proposes an MFT-enhanced WBC scheme for parallel-legged robots. 
Introducing the performance indices of MFT into a WBC is challenging due to the nonlinear relationship between MFT indices and the robot conﬁguration.
To overcome this challenge, we establish the MFT preferable space of the robot offline and formulate it as a polyhedron in the joint space at the acceleration level.
Then, the WBC employs the polyhedron as a soft constraint. As a result, the robot possesses high-speed and high-acceleration capabilities by satisfying this constraint.
The offline preprocessing relieves the online computation burden and helps the WBC achieve a 1kHz servo rate.
Finally, we validate the performance and robustness of the proposed method via simulations and experiments on a parallel-legged bipedal robot.

\end{abstract}

\section{Introduction}
\label{sec: INTRODUCTION}


A robot's performance index is a metric that can be used for synthesis, base placement and task planning applications. Recently, to achieve kinematically suitable configurations and highly dynamic task execution, the performance index plays an important role in the real-time control of robots \cite{zhang2016qp}, \cite{jin2017manipulability}, \cite{jin2020perturbed}, \cite{bellicoso2016perception}. 
For the redundant manipulator control, Zhang et al. proposed a quadratic
programming (QP) based scheme, which expressed the manipulability-maximizing criteria as a part of QP's cost function \cite{zhang2016qp}.
Jin et al. improved the computational efficiency of QP-based manipulability optimization by combining dynamic neural
networks \cite{jin2017manipulability}, \cite{jin2020perturbed}. 
For avoiding singularity during quadrupedal locomotion, Bellicoso-Hutter formulated the manipulability measures as an optimization criterion in a whole-body control (WBC) framework \cite{bellicoso2016perception}. 
The above works used manipulability as a performance index and embedded the performance index into a QP-based control framework to improve the dynamic performance of serial manipulators and serial-legged robots. 
Inspired by these successful works, we propose a novel control scheme for parallel-legged robot locomotion in this paper. This scheme introduces performance indices that reflect the motion and force transmissibility of parallel robots, as soft inequality constraints to QP-based whole-body controller for the enhancement of dexterity and robustness.

\begin{figure}[tbp]
\vspace{+0.3\baselineskip}
\centering
    \subfigure[]{
    \includegraphics[width=2.8cm]{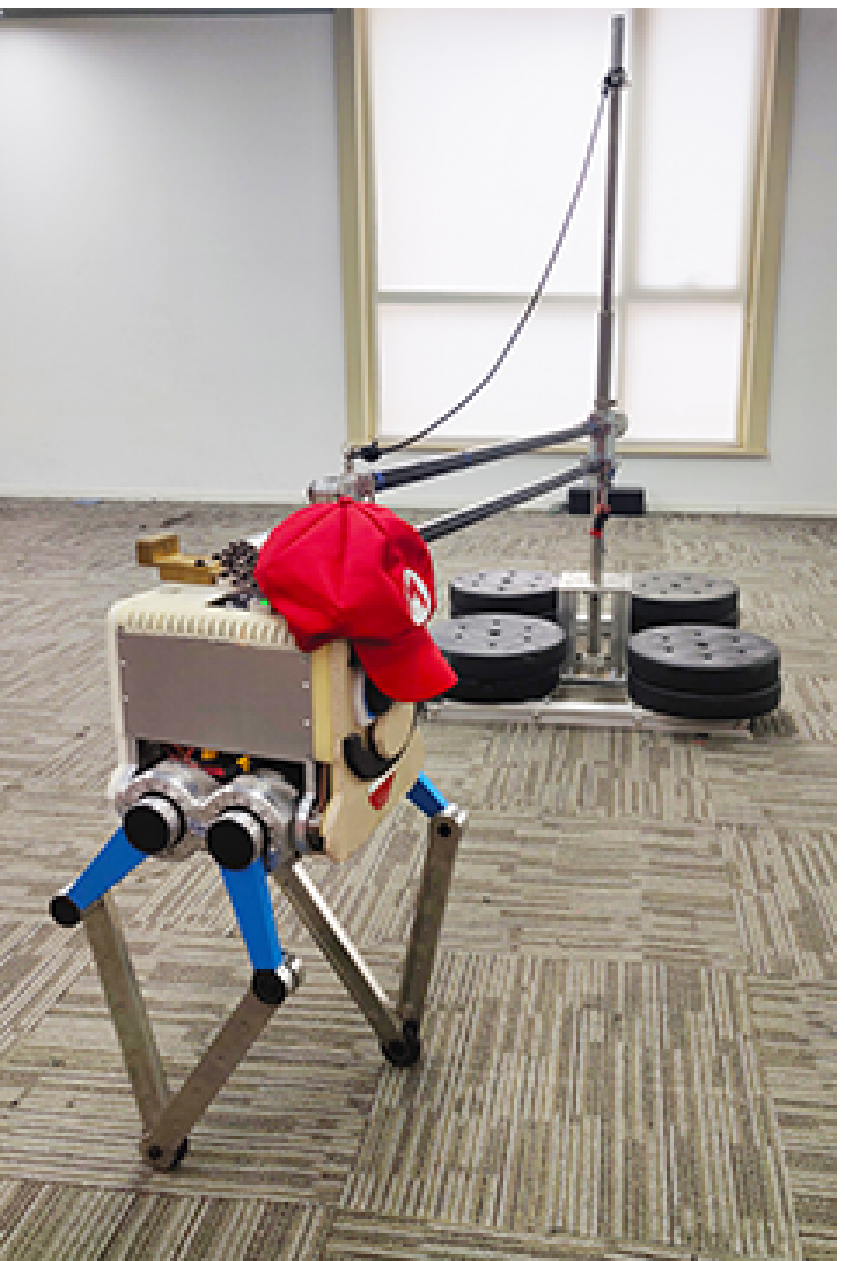}
    }
    \subfigure[]{
    \includegraphics[width=4.1cm]{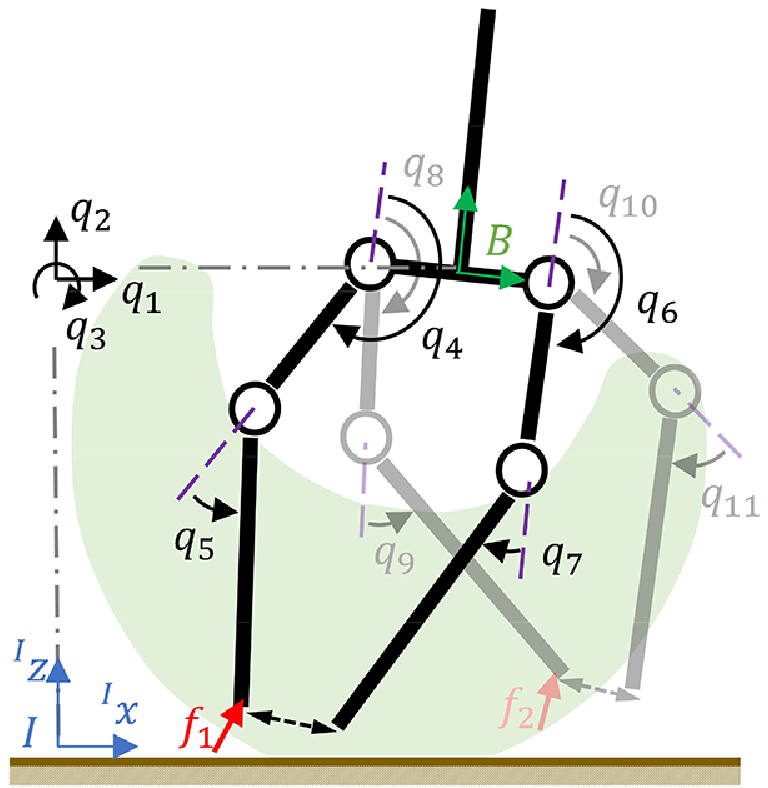}
    }
    \caption{(a) Parallel-legged bipedal robot. (b) The kinematic scheme of the robot. The dashed double-arrows at ankles represent the closed-loop holonomic constraints, and the light green area indicates the MFT preferable space of the right foot with respect to the floating-base.}
    \label{fig: robot-boom}
\end{figure}

The parallel mechanisms usually have low inertia, high stiffness, and more importantly, it presents the characteristics of high-load capacity, high-speed capability and excellent dexterity. Consequently, many legged robots use the parallel mechanism, e.g., ATRIAS \cite{hubicki2016atrias}, Cassie \cite{apgar2018fast}, Digit \cite{castillo2021robust}, Minitaur \cite{neil2014minitaur}, and Ollie \cite{wang2021balance}.
However, the controllers applied to these robots did not consider the performance index for parallel robots. 
The interaction between motion and force in parallel mechanisms has a signiﬁcant inﬂuence on the robot's kinematic and dynamic performance, e.g., speed ability, accuracy, acceleration capacity, and even power efﬁciency \cite{wang2010performance}. Therefore, the performance of motion/force transmissibility (MFT) was proposed to analyze this interaction, along with the indices used to evaluate the MFT \cite{takeda1995motion}.
Chen et al. proposed the power coefﬁcient to evaluate the MFT for single-loop mechanisms \cite{chen2007generalized}. The local transmission index (LTI) proposed in \cite{wang2010performance} and \cite{liu2012new} evaluated the MFT for multi-loop mechanism. Recently, the relationship between the MFT and the dynamic performance was analyzed, and the robot acceleration capacity index (RACI) was proposed for various parameters synchronous optimization \cite{liu2018novel}.
Compared to Jacobian-based performance indices (such as manipulability \cite{yoshikawa1985manipulability} and conditioning index \cite{gosselin1989optimum}), the MFT indices can better reflect the essential properties of parallel robots. In addition, for parallel robots with combined translational and rotational degrees of freedom, the MFT indices do not have the problem of physical inconsistency that the Jacobin-based indices suffer \cite{bowling2005dynamic}, \cite{merlet2006jacobian}.
Consequently, we employ MFT indices LTI and RACI to enhance the performance of the real-time control of parallel-legged locomotion.

Since the QP-based WBC method can efficiently cope with the limitations of actuators, the changing unilateral contact constraints and possibly conﬂicting tasks, it has been increasingly applied to the locomotion of legged robots 
\cite{dai2014whole}, \cite{feng2015optimization}, \cite{herzog2016momentum}, \cite{bellicoso2017dynamic}, \cite{morlando2021whole}.
Furthermore, for parallel-legged robots, this method can appropriately model the dynamics and kinematics of the closed-loop chains \cite{apgar2018fast}, \cite{kim2018computationally}, \cite{reher2020inverse}.
However, to the best of our knowledge, no prior work has addressed the MFT with WBC scheme in parallel-legged locomotion.
Since MFT indices have nonlinear relationships with the configurations of the robot, it is hard to express MFT indices into a QP and to achieve timely reactive MFT optimization in highly dynamic situations.
To overcome this challenge, we first establish the MFT preferable space of the robot offline and formulate it as a polyhedron in the joint space at the acceleration level. Then, a QP-based WBC scheme employs the polyhedron as a soft constraint to enforce MFT enhancement. Finally, we test our scheme on the dynamic locomotion of a parallel-legged biped (Fig.~\ref{fig: robot-boom}). The weight of the robot is 23kg. We deliberately install four heavy calves on the robot so that the mass ratio between the legs and the torso is about 40\%, which is not negligible according to \cite{kim2018control}. 
The results show that, with the proposed WBC scheme, the robot possesses high-speed and high-acceleration capabilities and achieve perception-less robust locomotion with external disturbances.

In summary, the main contribution is the proposed MFT-enhanced WBC scheme, referred to as MFT-WBC. This scheme combines offline discretization, polyhedral approximation and online QP optimization for introducing the MFT performance indices into a QP-based WBC framework. It can achieve high-speed online WBC computation, and improve the robot's reactive robustness to external impact and perception-less adaptation to uneven terrain. 
In the rest of the paper, Sec.~\ref{sec: INDICES OF MFT} introduces two important indices of MFT. Sec.~\ref{sec: WHOLE-BODY CONTROL SCHEME} states the implementation of the MFT-WBC scheme. Sec.~\ref{sec: SIMULATIONS AND EXPERIMENTS} demonstrates the simulation and experimental results. Finally, Sec.~\ref{sec: CONCLUSIONS} concludes this paper.

\section{Indices of Motion/Force Transmissibility }
\label{sec: INDICES OF MFT}

This section introduces two performance indices used to evaluate MFT briefly. 
First, the transmission angle $\alpha$ is used to measure the performance of single closed-loop mechanisms \cite{takeda1995motion}. It has a relationship with the power efficiency of the transmission force $\bf{f }$ and the speed $\bf{v }$ at the output end. The power efficiency is the ratio of the instantaneous power and its potential maximum value:
\begin{equation} \label{equ: transmission angle}
    \sin(\alpha) = {\frac{{\left| {{{\bf{f }}} \cdot {{\bf{v }}}} \right|}}{{\left\| {{\bf{f }}} \right\|}{\left\| {{\bf{v }}} \right\|}}},
\end{equation}
where ${{\left| {{{\bf{f }}} \cdot {{\bf{v }}}} \right|}}$ is the absolute value of the instantaneous power of ${\bf{f }}$ and ${\bf{v }}$, and $\left\| {} \right\|$ denotes the operator norm.
Liu \cite{wang2010performance} extended the theory of transmission angle and applied the power efficiency to the MFT evaluation of parallel mechanisms. \cite{wang2010performance} proposed the method to identify the transmission force (wrench) and permitted motion (twist) of parallel mechanisms using the screw theory. Further, the instantaneous power of the transmission force and permitted motion in a limb was formulated as: 
\begin{equation} \label{equ: instantaneous power}
    {\mathcal{P}}_{I} = {{\left| {{{{\bf{{\textit{\$}} }}_{T}}} \circ {{\bf{{\textit{\$}} }}_{I}}} \right|}},\quad {\mathcal{P}}_{O} = {{\left| {{{{\bf{{\textit{\$}} }}_{T}}} \circ {{\bf{{\textit{\$}} }}_{O}}} \right|}},
\end{equation}
where $\circ$ denotes reciprocal product in the screw theory, ${{{\bf{{\textit{\$}} }}_{T}}}$ denotes the transmission wrench, ${{{\bf{{\textit{\$}} }}_{I}}}$ and ${{{\bf{{\textit{\$}} }}_{O}}}$ denote the permitted twists at the input and output of the limb, respectively. The terms ${\mathcal{P}}_{I}$, ${\mathcal{P}}_{O}$ denote the instantaneous power at the input and output of the limb, respectively.
Similar to (\ref{equ: transmission angle}), by calculating the power efficiency, the input and output transmission indices were proposed as \cite{wang2010performance}:
\begin{subequations} 
\label{equ: equ: LTI I and O}
    \begin{align}
        &{\gamma _I} = \min \left\{ {\frac{{{{\mathcal{P}}_{I1}}}}{{{{\mathcal{P}}_{I1}}_{\max }}},\frac{{{{\mathcal{P}}_{I2}}}}{{{{\mathcal{P}}_{I2}}_{\max }}}, \cdots ,\frac{{{{\mathcal{P}}_{I{n_l}}}}}{{{{\mathcal{P}}_{I{n_l}}}_{\max }}}} \right\},\\
        &{\gamma _O} = \min \left\{ {\frac{{{{\mathcal{P}}_{O1}}}}{{{{\mathcal{P}}_{O1}}_{\max }}},\frac{{{{\mathcal{P}}_{O2}}}}{{{{\mathcal{P}}_{O2}}_{\max }}}, \cdots ,\frac{{{{\mathcal{P}}_{O{n_l}}}}}{{{{\mathcal{P}}_{O{n_l}}}_{\max }}}} \right\},
    \end{align}
\end{subequations}
where $n_l$ is the number of the limbs. ${\gamma _I}$ reflects the ability of transmitting motion and ${\gamma _O}$ reflects the ability of resisting external forces. 
The local transmission index
(LTI) was obtained based on the minimum value of ${\gamma _I}$ and ${\gamma _O}$ \cite{wang2010performance}:
\begin{equation} 
\label{equ: equ: LTI}
    {\gamma _{LTI}}{\rm{ = min}}\left\{ {{\gamma _I}{\rm{,}}{\gamma _O}} \right\}.
\end{equation}
The index ${\gamma _{LTI}} \in $ [0, 1] is dimensionless. Larger ${\gamma _{LTI}}$ means better MFT. ${\gamma _{LTI}} = 0$ indicates that the robot has input or output singularity.

In addition, 
in order to ensure the high-acceleration capability of the robot,
we also consider the robot acceleration capacity index (RACI) \cite{liu2018novel}. the MFT index RACI represents the acceleration capacity of the parallel robot with a given configuration and is expressed as:
\begin{equation} \label{equ: RACI}
    {\gamma _{RACI}} = \max \left\{ {{{\tilde \tau }_1},{{\tilde \tau }_2}, \cdots ,{{\tilde \tau }_{{n_l}}}} \right\},
\end{equation}
where ${{\tilde \tau}}$ is the normalized joint torque in the worst case with the end-effector's desired maximum translational and rotational accelerations.
The term ${{\tilde \tau}}$ is a nonlinear function of ${\mathcal{P}}_{I}$,  ${\mathcal{P}}_{O}$, the inertial parameters of the robot, and the specifications of the actuators. Refer to \cite{liu2018novel} for its detailed derivation. 
The value of ${{\tilde \tau}}$ is related to the abilities of transmitting motion and resisting external forces. 
A configuration with greater ${\gamma _{RACI}}$ indicates the robot requires larger joint torques to achieve the desired end-effector acceleration.

We use LTI and RACI as the MFT indices in our application, and introduce them to a QP-based WBC scheme.

\section{Implementation of MFT-enhanced Whole-Body Control Scheme}
\label{sec: WHOLE-BODY CONTROL SCHEME}

\begin{figure}[tb]
\vspace{+0.3\baselineskip}
    \centering
    \includegraphics[width=0.48\textwidth]{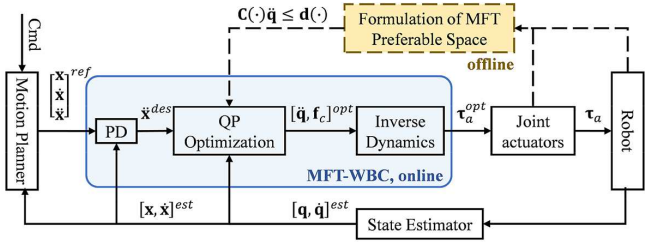}
    \caption{The locomotion control architecture. The solid lines and boxes denote the online parts, and the dashed ones denote the offline parts.}
    \label{fig: architecture}
\end{figure}

Our locomotion control architecture 
is shown in Fig.~\ref{fig: architecture}.
First, a state estimator provides the current robot states ${{\bf{x}}^{est}}$.
Given the user command, a motion planner generates the reference trajectories ${{\bf{x}}^{ref}}$ according to the robot states. Then the MFT-WBC generates optimized joint torques $\boldsymbol{\tau}_a^{opt}$ to track these trajectories and comply with specific physical constraints. 
In this paper,
we focus on the offline MFT formulation and the online QP-based WBC. 

\subsection{Dynamic Consistency of the Parallel-Legged Robot}
\label{sec: Dynamic Consistency of the Parallel-Legged Robot}

As shown in Fig.~\ref{fig: robot-boom}, with the torso constrained by a boom, the bipedal robot can move and pitch within the sagittal plane. Each leg of the robot 
is composed of a five-bar linkage mechanism,
which is actuated by two identical series elastic actuators (SEAs) at fore and rear hip joints. We model the robot as a rigid body system consisting of 11 generalized coordinates ${\bf{q}} \in {\mathbb{R}^{11}}$, 2 closed-loop holonomic constraints at passive ankles, and 4 contact forces of feet ${{\bf{f}}_c} = {[f_1^x,f_1^z,f_2^x,f_2^z]^T}$. In more detail,  the generalized coordinates ${\bf{q}}$ consist of the position and orientation of the floating base ${[{q_1},{q_2},q_{3}]^T}$, the position of the actuated ${[{q_4},{q_6},q_{8},q_{10}]^T}$ and the underactuated ${[{q_5},{q_7},q_{9},q_{11}]^T}$ joints.

The parallel-legged robot can be modelled as a rigid-body system with holonomic constraints caused by closed-loop chains, and the system dynamics can be written as:
\begin{equation} \label{equ: system EoM}
    \left[ {\begin{array}{*{20}{c}}
    {\bf{M}}&{ - {\bf{J}}_h^T}\\
    {{\bf{J}}_h}&{\bf{0}}
    \end{array}} \right]\left[ {\begin{array}{*{20}{c}}
    {{\bf{\ddot q}}}\\
    {{\bf{f}}_h}
    \end{array}} \right] = \left[ {\begin{array}{*{20}{c}}
    {{\bf{S}}_a^T{{\boldsymbol{\tau }}_a} + {\bf{J}}_c^T{{\bf{f}}_c} - {\bf{H}}}\\
    { - {{{\bf{\dot J}}}_h}{\bf{\dot q}}}
    \end{array}} \right],
\end{equation}
where $\bf{M}$ is the inertia matrix, $\bf{H}$ is the vector that accounts for Coriolis, centrifugal and gravitational forces. The vector ${\boldsymbol{\tau }}_a$ is the torque vector of actuated joints, and the matrix ${\bf{S}}_a$ 
is the actuation matrix selecting actuated degrees of freedom.
The vector of contact forces ${\bf{f}}_c$ is mapped to the generalized joint space through the combined contact Jacobian ${\bf{J}}_c$. A similar relationship can be found between the internal closed-loop holonomic constraint forces ${{\bf{f}}_h}$ and its corresponding Jacobian ${\bf{J}}_h$. 


By eliminating ${{\bf{f}}_h}$, an equation that fully describes the dynamics of the parallel-legged robot can be obtained as:
\begin{equation} \label{equ: null space EoM} 
    {\bf{M\ddot q}} + {\bf{N}}_h^T {{\bf{H}}} + {\bf{J}}_h^T{{\bf{\Lambda }}_h}{{{\bf{\dot J}}}_h}{\bf{\dot q}} - {\bf{N}}_h^T{\bf{S}}_a^T{{\boldsymbol{\tau }}_a} = {\bf{N}}_h^T{\bf{J}}_c^T{{\bf{f}}_c},
\end{equation}
where,
${{\bf{\Lambda }}_h} = {({{\bf{J}}_h}{{\bf{M}}^{ - 1}}{\bf{J}}_h^T)^{ - 1}}$, ${\bf{J}}_h^\#  = {{\bf{M}}^{ - 1}}{\bf{J}}_h^T{{\bf{\Lambda }}_h}$, and ${{\bf{N}}_h} = {\bf{I}} - {\bf{J}}_h^\# {{\bf{J}}_h}$
denote the apparent inertia, dynamically consistent inverse, and null space projector of the holonomic constraint, respectively.
The floating-base dynamics can be derived from (\ref{equ: null space EoM}) with the floating-base selection matrix ${{\bf{S}}_f}$:
\begin{equation} \label{equ: Dynamic consistency} 
    {{\bf{S}}_f}({\bf{M\ddot q}} + {\bf{N}}_h^T {{\bf{H}}}) = {{\bf{S}}_f}{\bf{N}}_h^T{\bf{J}}_c^T{{\bf{f}}_c},
\end{equation}
and the last two terms in the left-hand-side of (\ref{equ: null space EoM}) is canceled, since the internal forces do not influence the centroidal dynamics \cite{orin2013centroidal}.
The actuated joint torques can be obtained by inverse dynamics:
\begin{equation} \label{equ: torque actuated}
    \begin{split}
    {{\boldsymbol{\tau }}_a} = 
    &{\left( {{{\bf{S}}_a}{\bf{N}}_h^T{\bf{S}}_a^T} \right)^{ - 1}}{{\bf{S}}_a}  \\
    &\left( {{\bf{M\ddot q}} + {\bf{N}}_h^T{\bf{H}} + {\bf{J}}_h^T{{\bf{\Lambda }}_h}{{{\bf{\dot J}}}_h}{\bf{\dot q}} - {\bf{N}}_h^T{\bf{J}}_c^T{{\bf{f}}_c}} \right).
    \end{split}
\end{equation}

\subsection{Formulation of MFT Preferable Space}
\label{sec: Formulation of MFT Preferable Space}

Our approach to enhancing MFT in a control scheme is to restrict the robot within a workspace that possesses promising MFT, resulting in the following definition.

\vspace{+0.3\baselineskip}
\noindent{\bf{Definition 1}}
(MFT preferable space)\textbf{.}
\textit{Let $\boldsymbol{\gamma}$ to denote a set of MFT indices, and $[{\underline{\boldsymbol{\gamma}}},{\overline{\boldsymbol{\gamma}}}]$ denotes an user-defined reasonable range of these indices. A workspace ${{}^{\boldsymbol{\gamma}}\bf{Q}}$ of a configuration of a parallel mechanism is the MFT preferable space with respect to $\boldsymbol{\gamma}$, if and only if, the value of the MFT indices of all points in ${{}^{\boldsymbol{\gamma}}\bf{Q}}$ are within $[{\underline{\boldsymbol{\gamma}}},{\overline{\boldsymbol{\gamma}}}]$.}
\vspace{+0.3\baselineskip}

It is challenging to consider MFT indices as an optimization
criterion or a constraint in QP since MFT indices have nonlinear relationships with the robot’s configuration. To overcome this challenge,
we first approximate the robot's MFT preferable space as polyhedra, then convert the resulting polyhedra into an augmented polyhedron in the generalized joint space at the acceleration level.

\begin{figure}[btp]
    \vspace{+0.3\baselineskip}
    \centering
    \includegraphics[width=8.6cm]{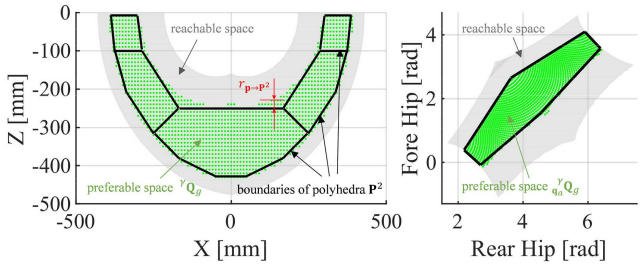}
    \caption{The MFT preferable space of each leg in the Cartesian space (left) and the actuated joint space (right). The light gray area denotes its reachable space, and the gridded green area denotes its gridded MFT preferable space. The boundaries of polyhedra are marked by black solid lines. the \textit{Rear Hip} and \textit{Fore Hip} correspond to ${q_4}$ and ${q_6}$ of the right leg or ${q_8}$ and ${q_{10}}$ of the left leg in Fig.~\ref{fig: robot-boom}. }
    \vspace{-0\baselineskip}
    \label{fig: MFT Preferable Space}
\end{figure}

For each parallel mechanism in a parallel-legged robot, 
we first discretize the reachable space of its end effector in the Cartesian space with a user-defined resolution. Then, we get the gridded MFT preferable space ${{}^{\boldsymbol{\gamma}}\bf{Q}}_g$ by traversing both the former discretized reachable space and the given indices set $\boldsymbol{\gamma}$, according to Definition 1.
Hereafter, we approximate ${{}^{\boldsymbol{\gamma}}\bf{Q}}_g$ with several 3-dimensional polyhedra ${\bf{P}}^3$ conservatively using the method provided by \cite{lo2014finite}. This approximation fulfills the following conditions:
\begin{enumerate} 
  \item $\forall {\bf{p}} \in {\bf{P}}_g^3 : {\bf{p}} \in {{}^{\boldsymbol{\gamma}}\bf{Q}}_g$,
  \vspace{+0.2\baselineskip}
  \item $\forall {\bf{p}} \notin {\bf{P}}_g^3 \wedge {\bf{p}} \in {{}^{\boldsymbol{\gamma}}\bf{Q}}_g : r_{{\bf{p}} \to {\bf{P}}^3} \le {\underline{r}}$,
\end{enumerate}
where ${\bf{P}}_g^3$ is the gridded ${\bf{P}}^3$ with the same resolution as ${{}^{\boldsymbol{\gamma}}\bf{Q}}_g$, $r_{{\bf{p}} \to  {\bf{P}}^3 }$ denotes the distance from the point ${\bf{p}}$ to the boundary of ${\bf{P}}^3$ , and ${\underline{r}} \ge 0$ is a scalar reflecting the conservative degree of the approximation. The left-hand-side of Fig.~\ref{fig: MFT Preferable Space} shows the corresponding spaces of each parallel leg of our robot. The grid resolution is $10\times10$ mm. The indices LTI and RACI are used to evaluate the MFT. We set $0.7 \le {\gamma _{LTI}} \le 1$ based on prior knowledge, and set $0 \le {\gamma _{RACI}} \le 174.54$ $({\rm{N/}}\sqrt{\rm{kg}})$ based on the specifications of the actuators, so that a maximum acceleration 60 ${\rm{m/}}{{\rm{s}}^2}$ at the foot can be achieved during locomotion. 
Since the robot is a planar robot, the dimension of the polyhedra become 2.

The polyhedra ${\bf{P}}^3$ can be expressed as the intersection of several halfspaces in linear inequalities. For the case shown in the left-hand-side of Fig.~\ref{fig: MFT Preferable Space}, the linear inequalities are piecewise.
By collecting the MFT preferable polyhedra of all the legs of the robot, we can obtain an augmented MFT preferable polyhedron in the workspace:
\begin{equation} \label{equ: polyhedron in p}
    {{\bf{P}}_{\bf{p}}^m} = \{ {{\bf{p}} \in {\mathbb{R}^m}}|{\bf{{A}p}} \le {\bf{b}}\}, 
\end{equation}
where the subscript ${\bf{p}}$ is used to identify the axes of the space in which the polyhedron resides.
${\bf{A}} \in {\mathbb{R}^{{n_h} \times m}},{\bf{b}} \in {\mathbb{R}^{{n_h}}}$, ${n_h}$ is the number of halfspaces in the polyhedron, $m = 3n_s$ denotes the dimension of the polyhedron, and $n_s$ is the number of parallel mechanisms.

Since the generalized acceleration vector ${\bf{\ddot q}}$ is one of the optimization variables, 
the polyhedron ${{\bf{P}}_{\bf{p}}^m}$ should be converted to a polyhedron in the generalized joint space at the acceleration level:
\begin{equation} \label{equ: polyhedron in qddot}
    {\bf{P}}_{{\bf{\ddot q}}}^n = \{ {\bf{\ddot q}} \in {\mathbb{R}^n}|{\bf{C\ddot q}} \le {\bf{d}}\}, 
\end{equation}
where ${\bf{C}} \in {\mathbb{R}^{{n_h} \times n}},{\bf{d }} \in {\mathbb{R}^{{n_h}}}$, $n$ is the dimension of the generalized joint space.
This conversion can be conducted by approximating ${\bf{p}}$ with its Taylor expansion around the current instant $\hat t$. For the $k^{\rm{th}}$ element of $\bf{p}$, the position after an interval $\Delta t$ from $\hat t$ can be approximated as:
\begin{equation} \label{equ: Taylor for p}
    \begin{array}{ll}
    {{{p}}_k}&\cong {{\hat p}_k} + {{\hat {\dot p}}_k} + \frac{1}{2}{{\hat {\ddot p}}_k}\Delta t^2\\
    &  = {{\hat p}_k} + {{\bf{J}}_{(k,:)}}{\bf{\hat {\dot q}}}\Delta t + \frac{1}{2}\left( {{{\bf{J}}_{(k,:)}}{{\bf{\hat{\bf{\ddot q}}}}} + {{[{\bf{\dot J\dot q}}]}_k}} \right)\Delta t^2,
    \end{array}
\end{equation}
where the terms with hat represent the values at $\hat t$, the subscript ${(k,:)}$ denotes the $k^{\rm{th}}$ row in a matrix. The terms ${\bf{J}}$ and ${\bf{\dot J\dot q}}$ satisfy 
${\bf{\dot p}} = {\bf{J\dot q}}$ and ${\bf{\ddot p}} = {\bf{J\ddot q}} + {\bf{\dot J\dot q}}$, where $\bf{J}$ is the Jacobian of the end effector.
Substituting (\ref{equ: Taylor for p}) into (\ref{equ: polyhedron in p}), the polyhedron defined in (\ref{equ: polyhedron in qddot}) can be then obtained as:
\begin{equation} \label{equ: C}
    {{\bf{C}}_{(i,j)}} = \frac{1}{2}\Delta {t^2}\sum\limits_{k = 1}^m {\left( {{{\bf{A}}_{(i,k)}}{{\bf{J}}_{(k,j)}}} \right)},
\end{equation}
and
\begin{equation} \label{equ: d}
    {{\bf{d}}_i} = {{\bf{b}}_i} - \sum\limits_{k = 1}^m {\left( {{{\bf{A}}_{\left( {i,k} \right)}}\left( {{{{\bf{\hat{p}}}}_k} + {{\bf{J}}_{(k,:)}}{\bf{\hat {\dot{q}}}}\Delta t + \frac{1}{2}{{[{\bf{\dot{J}}}{\bf{ {\dot{q}}}}]}_k}\Delta {t^2}} \right)} \right)},
\end{equation}
where the subscript ${(i,j)}$ denotes the entry of a matrix.

As shown in Fig.~\ref{fig: architecture}, the offline formulation of MFT preferable space provides the structure of ${\bf{C}}( \cdot )$ and ${\bf{d}}( \cdot )$. These operations allow MFT to be introduced into the QP framework and greatly relieve the burden of online computation.

In our case, for the planar robot in Fig.~\ref{fig: robot-boom}, 
we can conduct polyhedron approximation in its actuated joint space through the bijective inverse kinematics, as shown in the right-hand-side of Fig.~\ref{fig: MFT Preferable Space}. 
The gridded MFT preferable space ${}_{{{\bf{q}}_a}}^{\boldsymbol{\gamma }}{{\bf{Q}}_g}$ can be approximated by a hexagon.
We express the robot's MFT preferable polyhedron in the actuated joint space as 
${\bf{P}}_{{\bf{q}}_a}^4 = \{ {{\bf{q}}_a} \in {{\mathbb{R}}^4}|{\bf{A}}{{\bf{q}}_a} \le {\bf{b}}\}$, where ${{\bf{q}}_a} = {[{q_4},{q_6},{q_8},{q_{10}}]^T}$, ${\bf{A}} \in {\mathbb{R}^{{12} \times 4}},{\bf{b }} \in {\mathbb{R}^{{12}}}$. Then we obtain the polyhedron
${\bf{P}}_{{\bf{\ddot q}}}^{11} = \{ {\bf{\ddot q}} \in {\mathbb{R}^{11}}|{\bf{C\ddot q}} \le {\bf{d}}\}$ with (\ref{equ: C}) and (\ref{equ: d}) by replacing ${{\bf{\hat p}}}$ and ${\bf{J}}$ with ${\bf{\hat q}}_a$ and ${\bf{S}}_a$, respectively. 
In this way, we have ${\bf{C}} \in {\mathbb{R}^{{12} \times 11}},{\bf{d }} \in {\mathbb{R}^{{12}}}$. The number of linear inequalities is reduced, and piecewise processing is eliminated. 

\subsection{QP Problem for Parallel-Legged Robot Locomotion}
\label{sec: QP Problem for Parallel-Legged Robot Locomotion}

The online MFT-enhanced WBC scheme for parallel-legged robot locomotion is formulated as follows.

\vspace{+0.4\baselineskip}
\noindent{\bf{Problem 1.}}
\begin{subequations} \label{equ: cost function}
    \begin{align}
        &\mathop {\min }\limits_{{\bf{\ddot q}},{{\bf{f}}_c},{\boldsymbol{\epsilon }}}
        \left\| {{{\bf{J}}_x }{\bf{\ddot q}} + {{{\bf{\dot J}}}_x }{\bf{\dot q}} - {{{\bf{\ddot x }}}^{des}}} \right\|_{{{\bf{W}}_x }}^2 + \\
        &\quad \quad 
        \left\| {{{\bf{f}}_c}} \right\|_{{{\bf{W}}_{{f_c}}}}^2 + \left\| {{{\bf{f}}_c} - {\bf{f}}_c^ - } \right\|_{{{\bf{W}}_{\delta {f_c}}}}^2 + \\
        &\quad \quad 
        \textcolor{blue}{\left\| {\boldsymbol{\epsilon }} \right\|_{{{\bf{W}}_\varepsilon }}^2}
    \end{align}
\end{subequations}
\noindent{\textit{subject to}}
\begin{subequations} \label{equ: constraint function}
    \begin{align}
        &{{\bf{S}}_f}({\bf{M\ddot q}} + {\bf{N}}_h^T {{\bf{H}}}) = {{\bf{S}}_f}{\bf{N}}_h^T{\bf{J}}_c^T{{\bf{f}}_c} \\
        &{\bf{U}}{{\bf{f}}_c} \le {\bf{0}}\\
        &{{\boldsymbol{\tau }}_a^{\min }} \le {{\boldsymbol{\tau }}_a} \le {{\boldsymbol{\tau }}_a^{\max }} \\
        &{{\bf{\ddot q}}^{\min }} \le {\bf{\ddot q}} \le {{\bf{\ddot q}}^{\max }} \\
        &\textcolor{blue}{{\bf{C\ddot q}} \le {\bf{d}} + {\boldsymbol{\epsilon }}},
    \end{align}
\end{subequations}
\noindent{\textit{where}}
\begin{equation} \label{equ: PD controller}
    {{{\bf{\ddot x}}}^{des}} = {{{\bf{\ddot x}}}^{ref}} + {{\bf{K}}_p}\left( {{{\bf{x}}^{ref}} - {{\bf{x}}^{est}}} \right) + {{\bf{K}}_d}\left( {{{{\bf{\dot x}}}^{ref}} - {{{\bf{\dot x}}}^{est}}} \right),
\end{equation}
${{\bf{\ddot x}}}^{des}$ denotes the desired task accelerations,
${{\bf{K}}_p}$ and ${{\bf{K}}_d}$ are diagonal feedback matrices, ${{\bf{J}}_{x} }$ is the task space Jacobian, ${\boldsymbol{\epsilon }}$ is a relax variable and will be explained in detail later.

We get the desired task accelerations of the floating-base, the swing foot and the support foot through a PD controller (\ref{equ: PD controller}) in operational space, as shown in Fig.~\ref{fig: architecture}. 
Then, we minimize the weighted Euclidean distance of the operational space trajectory tracking error in (\ref{equ: cost function}a).
Working as a regularization to ${\bf{f}}_c$, the first term of (\ref{equ: cost function}b) aims to reduce the risk of slippage in case of model uncertainty \cite{bellicoso2016perception}. To avoid jumps in the actuation signals when a foot is touching down or lifting off, we penalize changes in contact forces by the second term of (\ref{equ: cost function}b) \cite{feng2015optimization}.
Constraint (\ref{equ: constraint function}a) identical to (\ref{equ: Dynamic consistency}) enforces dynamic consistency. The contact forces' bounds and friction constraints are enforced through (\ref{equ: constraint function}b) compactly, in which the friction cones are approximated as pyramids \cite{grizzle2014models}. 
The torque and joint limits are enforced in the inequality constraints (\ref{equ: constraint function}c) and (\ref{equ: constraint function}d), respectively.

The main distinctive feature of our scheme is that Problem 1 employs the polyhedron of MTF preferable space as (\ref{equ: cost function}c) and (\ref{equ: constraint function}e) to enhance the MFT performance.
We introduce the relax variable ${\boldsymbol{\epsilon }}$ for the MFT preferable polyhedron so it can be easily implemented as a soft constraint to the QP-based WBC method. Furthermore, we can tune its corresponding weight with respect to the convenient execution of tasks.
The relax variable ${\boldsymbol{\epsilon }}$ has the same dimension as ${\bf{d}}$. The values of ${\bf{C}}$ and ${\bf{d}}$ are updated online with (\ref{equ: C}) and (\ref{equ: d}) according to the current states ${\bf{\hat {p}}}$, ${\bf{\hat {q}}}$, ${\bf{\hat {\dot {q}}}}$ and the time interval $\Delta t$.
Here, $\Delta t$ is the prediction length in (\ref{equ: Taylor for p}), which makes the controller have the effect of predictive control.



At this point, the proposed MFT-WBC is fully specified. With the optimization variables ${{{[{{{\bf{\ddot q}}}^T},{\bf{f}}_c^T,{{\boldsymbol{\epsilon }}^T}]}^T}} \in {{\mathbb{R}}^{27}}$, Problem 1 uses weight matrices ${{\bf{W}}_ * }$ to embody the relative priority among tasks $*$. Solving the QP and inverse dynamics yields the control command ${\boldsymbol{\tau }}_a^{opt}$.

\section{Simulations and Experiments}
\label{sec: SIMULATIONS AND EXPERIMENTS}


\subsection{Implementation Setup}
\label{sec: Implementation Setup}

\subsubsection{Motion Planner Setup}
\label{sec: Motion Planner}

The user-defined height of the floating-base $B$ and the desired average sagittal speed (i.e., $[{}^Iz_B^{cmd},\overline{v}^{cmd}]^T$) are the inputs of the motion planner. The planner outputs the trajectories of the floating-base and two feet for the ground frame \textit{I} according to robot states.
Due to the underactuated nature of the bipedal walking, only the height and pitch of the floating-base are planned, leaving its sagittal movement as a passive dynamic process:
\begin{equation}    \label{equ: floating base plan}
    {}^Iz_B^{ref} = {}^Iz_B^{cmd},\quad {}^I\theta _B^{ref} = 0,
\end{equation}
The trajectory of the support foot is set to its current location.
The swing foot planning aims to regulate the average sagittal speed of the floating-base towards $\overline{v}^{cmd}$. We plan the target pre-impact foot placement with a discrete P-type controller: 
\begin{equation}    \label{equ: step control}
    {}^Ix_{sw}^{ - *} = {}^Ix_C^ -  + {k_v}{}^I\dot x_C^ -  + {k_p}({}^I\dot x_C^ -  - {}^I\dot x_C^{ -  * })
\end{equation}
where ${[{}^Ix_C^ - ,{}^I\dot x_C^ - ]^T}$ is the predicted horizontal pre-impact center of mass (CoM) state \cite{guo2021fast}, ${}^I\dot x_C^{ -  * }$ is the desired pre-impact CoM speed determined by $\overline{v}^{cmd}$ \cite{xiong2019orbit}. We set $k_v$ = 0.16, $k_p$ = 0.08 in this application. The nominal step duration of the walking gait is 0.35 sec. 
The cubic polynomial trajectory of the swing foot is generated from its lift-off position to the target pre-impact position smoothly \cite{apgar2018fast}. All these trajectories together with (\ref{equ: floating base plan}) will be tracked through WBC with (\ref{equ: cost function}a) and (\ref{equ: PD controller}).

\subsubsection{MFT-WBC Setup}
\label{sec: MFT-WBC Setup}

The WBC utilizes RBDL \cite{felis2017rbdl} for rigid body dynamics and qpOASES \cite{ferreau2014qpoases} as the QP solver with the option \textit{hotstart}, running at 1 kHz.
Since the SEA shows an average settling time of ${{\bar t}_s}$ = 0.025 sec for step torque signals, we set $\Delta t$ = 2${{\bar t}_s}$ for the prediction (\ref{equ: Taylor for p}) empirically.
The weights of tasks are specified in Table~\ref{tab: Weights of Tasks}, where ${w_{z}}$, ${w_{\theta}}$, ${w_{foot}^{sw}}$, and ${w_{foot}^{st}}$ denote the weights for tracking trajectories of floating-base, swing foot and support foot, respectively. And the terms ${w_{f_c}}$, ${w_{\delta f_c}}$ and ${w_{\epsilon}}$ are the diagonal elements of ${{\bf{W}}_{f_c}}$, and${{\bf{W}}_{\delta f_c}}$ and ${{\bf{W}}_{\epsilon}}$, respectively.
\begin{table}[tbp]
\vspace{+0.6\baselineskip}
\caption{Weights of Tasks for Problem 1}
\label{tab: Weights of Tasks}
    \begin{center}
        \begin{tabular}{cccccccc}
            \hline
            ${w_{z}}$ & ${w_{\theta}}$ & ${w_{foot}^{sw}}$ & ${w_{foot}^{st}}$ & ${w_{f_c}}$ & ${w_{\delta f_c}}$ & ${w_{\epsilon}}$ \\
            \hline
            1 & 1 & 1 & 1e2 & 1e-3 & 5e-2 & 1e3 \\
            \hline
        \end{tabular}
    \end{center}
\end{table}

\vspace{+0.3\baselineskip}
\noindent{\bf{Remark 1.}} 
\textit{In the push recovery and uneven terrain tests, the motion planner only stabilizes the sagittal speed of the floating-base by re-planning the swinging foot but does not react to other robot state changes or singularity avoidance, nor does it re-plan the trajectories of the floating-base or the support foot.}

\subsection{Simulation Results}
\label{sec: Simulation Results}

We conducted several push-recovery tests for the bipedal robot in simulation to verify the proposed method.
The simulation environment is the open-source robot simulator Webots.
Two different schemes were tested: One was the proposed MFT-WBC with the setup in Sec.~\ref{sec: Implementation Setup}; The other one, labeled SA-WBC, was the same as the former one, except that it removed the MFT terms (\ref{equ: cost function}c)(\ref{equ: constraint function}e) from Problem 1 and added an inequality constraint for singularity avoidance as follow:
\begin{equation} \label{equ: singularity avoidance}
    {\bf{q}}_{ua}^{\min } \le {{\bf{S}}_{ua}}\left( {{\bf{\hat q}} + {\bf{\hat {\dot q}}}\Delta t + \frac{1}{2}{\bf{\ddot q}}\Delta {t^2}} \right) \le {\bf{q}}_{ua}^{\max }
\end{equation}
where the terms with hat represent the values at the current instant, ${\bf{S}}_{ua}$ selects the underactuated joints ${{\bf{q}}_{ua}} = {[{q_5},{q_7},{q_9},{q_{11}}]^T}$. ${\bf{q}}_{ua}^{\min }$ and ${\bf{q}}_{ua}^{\max }$ denote the singularity boundaries. Constraint (\ref{equ: singularity avoidance}) is equivalent to constraining the leg in its reachable space, as shown in Fig.~\ref{fig: MFT Preferable Space}.

For every test, while stepping in place, the robot received a sagittal impulse when its one foot just lifted off. The push recovery of the robot at this moment is the most difficult since it takes about the entire swing duration to make a new touch down.
The impulse magnitude started at 3 Ns and increased in a step of 0.1 Ns until we found the maximum impulse $I_{max}$, from which the robot could recover.

Five groups of the above tests were conducted with different reference height of floating-base ${}^Iz_B^{ref}$, and the results are shown in Table~\ref{tab:maximum impulse}.
For the first three groups of tests, the heights ${}^Iz_B^{ref}$ were less than 0.43 m, which were relatively good plannings. This is because the robot's legs were naturally within their MFT preferable space when maintaining these floating-base heights (0.38 m, 0.40 m, 0.42 m) to step in place. On the contrary, the last two groups of tests had relatively bad plannings since the robot's support legs were always outside their MFT preferable space when stepping in place.

\begin{table}[tbp]
\vspace{+0.6\baselineskip}
\caption{Results of Push-Recovery Tests in Simulation}
\label{tab:maximum impulse}
    \begin{center}
        \begin{tabular}{cccccc}
            \hline
            ${}^Iz_B^{ref} $ [m] & 0.38 & 0.40 & 0.42 & 0.44 & 0.46 \\
            \hline
            $I_{max}$ of SA-WBC [Ns]& 10.4 & 9.7 & 9.1 & 7.0 & 4.5 \\
            $I_{max}$ of MFT-WBC [Ns]& 10.7 & 10.6 & 10.6 & 9.5 & 6.6 \\
            Percentage of Increase [\%]& 2.9 & 9.3 & 16.5 & 35.6 & 44.4 \\
            \hline
        \end{tabular}
    \end{center}
\end{table}

The results in Table~\ref{tab:maximum impulse} show that, compared with SA-WBC, the MFT-WBC can improve the robustness of the system. The maximum sustainable impulse increased 16.5\% for the relatively good plannings and 44.4\% for the relatively bad plannings. 
The proposed scheme can correct the trajectory to satisfy the MFT preferable space with a worse planning. Consequently, the performance as a percentage is increased significantly compared to SA-WBC.

\begin{figure}[tbp]
\vspace{+0.3\baselineskip}
\centering
    \subfigure[Trajectories of hip joint angles]{
    \includegraphics[width=8.6cm]{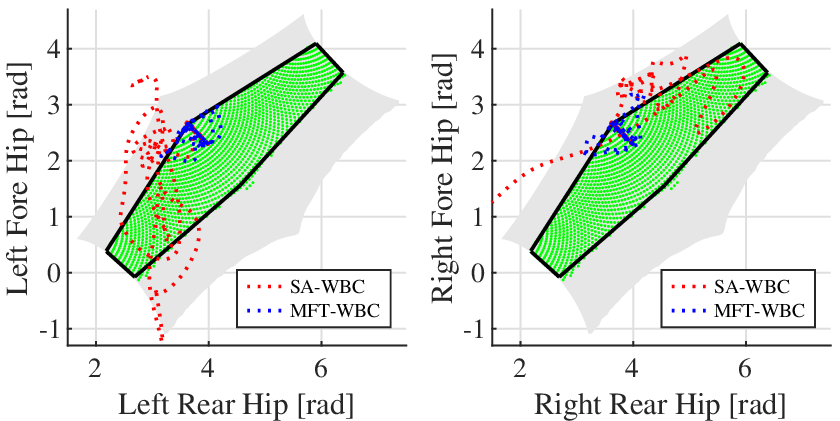}
    }
    \subfigure[State of the floating-base]{
    \includegraphics[width=8.6cm]{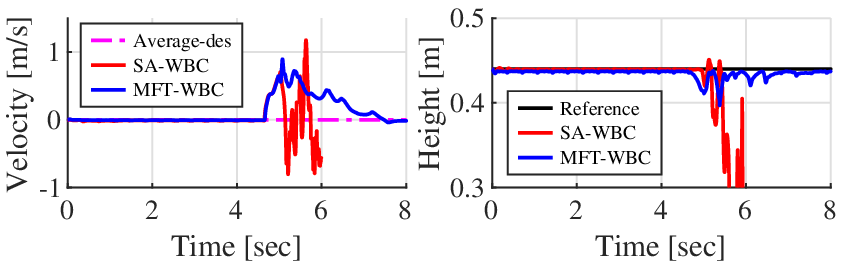}
    }
    \caption{Simulation results for the case ${}^Iz_B^{ref}=$ 0.44m, $I=$7.1Ns}
    \label{fig: Simulation results}
\end{figure}

For the case ${}^Iz_B^{ref}=$ 0.44 m, by applying $I=$ 7.1 Ns, 
the result trajectories of the hip joint angles are shown in Fig.~\ref{fig: Simulation results} (a), and the state of the floating-base is shown in Fig.~\ref{fig: Simulation results} (b). 
For the robot with MFT-WBC, when stepping without external impulse, the actual height of the floating-base was about 0.43 m, and the legs stayed within the MFT preferable space. After receiving the impulse at 4.66 sec, the MFT-WBC tried to keep the robot in the specified MFT preferable space by sacrificing the execution of other tasks, such as the height and attitude of the floating-base, resulting in that the robot retained the desired MFT and kept away from its singularity. Finally, the robot recovered from the external disturbance.
However, under the same external impulse, the SA-WBC diverged and the robot reached its singularity, although singularity avoidance constraint was considered through (\ref{equ: singularity avoidance}). 
If the leg is outside the MFT preferable space but within the reachable boundary, sustaining the same impulse requires significant joint torques that the actuators can not exert due to the poor MFT.
These results demonstrate the advantage of MFT-WBC in improving the robustness of the system, and similar results can be obtained from other groups of tests with different plannings and impulse magnitudes. 


\subsection{Experimental Results}
\label{sec: Experimental Results}

\begin{figure}[tbp]
\vspace{+0.3\baselineskip}
\centering
    \subfigure[]{
    \includegraphics[width=1.9cm]{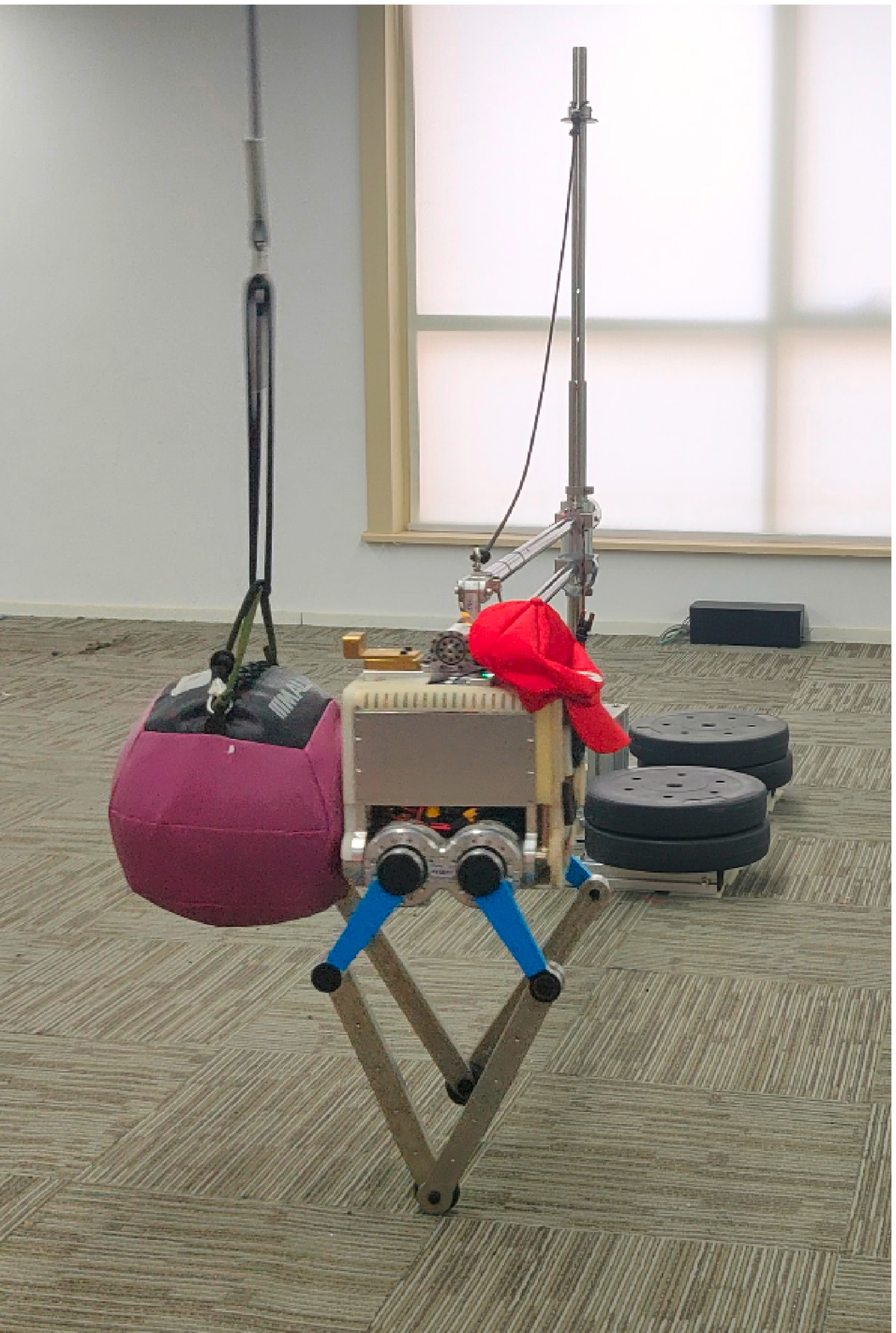}
    }
    \subfigure[]{
    \includegraphics[width=6.1cm]{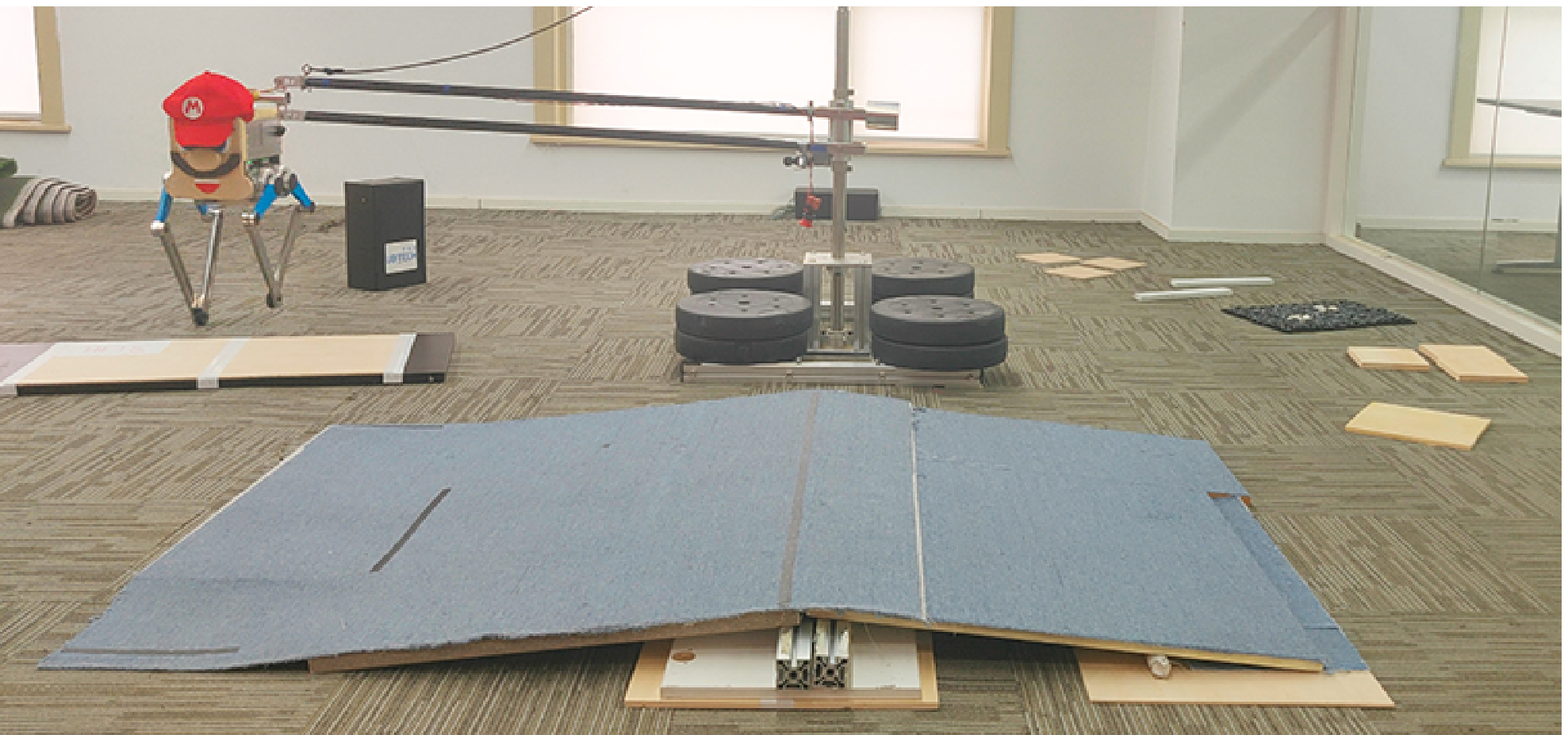}
    }
    \caption{(a) Push recovery. (b) Uneven terrain. In (b), the robot blindly walked through a 3 cm board, a ${6^ \circ }$ slope, as well as some rough terrains about 2 cm high composed of wooden/rubber boards and pebbles, tracking the specified average speed and upper body posture.}
    \label{fig: Experimental scene}
\end{figure}

As shown in Fig.~\ref{fig: Experimental scene}, we conducted two experiments on the robot in Fig.~\ref{fig: robot-boom} with the proposed MFT-WBC.

\subsubsection{Push Recovery}
\label{sec: Exp Push Recovery}

\begin{figure}[tbp]
\centering
    \subfigure[Sagittal velocity of the floating-base]{
    \includegraphics[width=8.6cm]{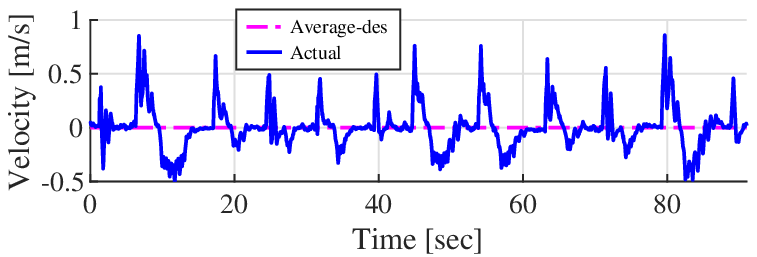}
    }
    \subfigure[State of the floating-base around the second push]{
    \includegraphics[width=8.6cm]{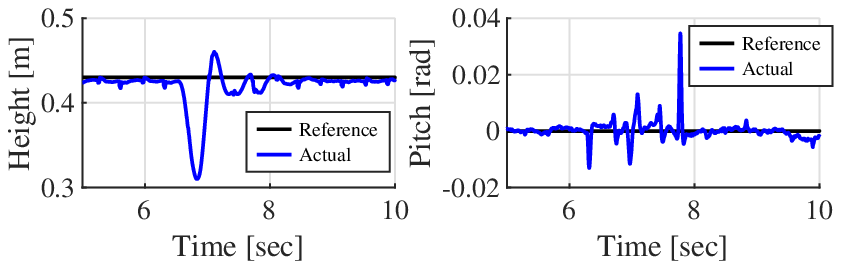}
    }
    \subfigure[Keyframes of the second push video clip]{
    \includegraphics[width=8.6cm]{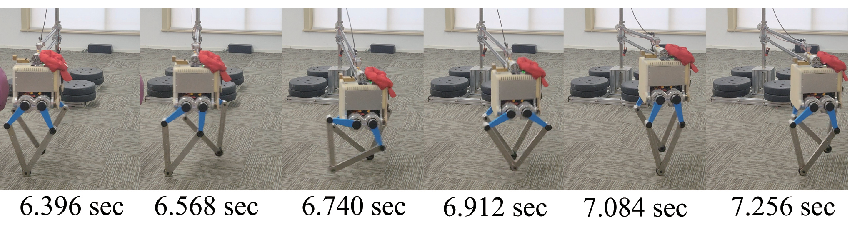}
    }
    \caption{Push recovery results}
    \label{fig: Exp push recovery results}
\end{figure}

Similar to the simulation, during stepping in place, the robot was hit by a 5 kg (22\% of robot's weight) wall ball in the sagittal direction 12 times. The sagittal velocity of the floating-base is shown in Fig.~\ref{fig: Exp push recovery results} (a), the robot recovered from instant velocity change up to 0.9 m/s. The state of the floating-base around the second push is presented in Fig.~\ref{fig: Exp push recovery results} (b), and the keyframes within 1 sec after the second push is shown in Fig.~\ref{fig: Exp push recovery results} (c). As we can see, the right leg was about to reach its singularity after 6.568 sec. 
In this situation, the MFT constraint lowered the height of the robot to keep the legs within their MFT preferable space. The MFT-WBC enhanced the robustness of the system by considering the MFT preferable space, which was essential to the recovery of the robot. Consequently, the execution of the latter tasks was ensured.

\subsubsection{Uneven Terrain}
\label{sec: Exp Uneven Terrain}

\begin{figure}[tbp]
\vspace{+0.3\baselineskip}
\centering
    \subfigure[State of the floating-base]{
    \includegraphics[width=8.6cm]{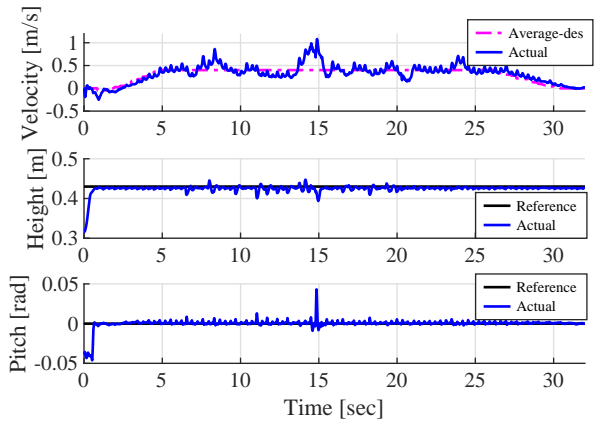}
    }
    \subfigure[State of left leg during up-slope]{
    \includegraphics[width=8.6cm]{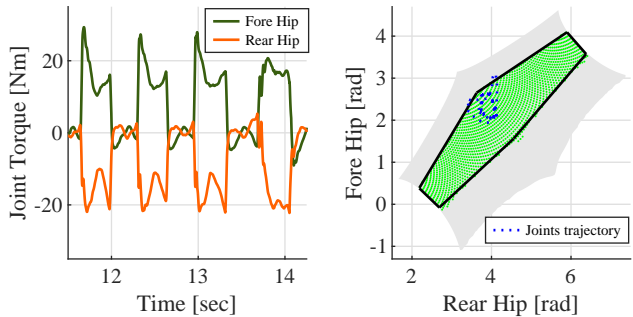}
    }
    \caption{Uneven terrain results}
    \label{fig: Exp uneven terrain results}
\end{figure}

The robot successfully passed the uneven terrain test in Fig.~\ref{fig: Experimental scene} (b). 
The experimental data collected during uneven terrain locomotion are shown in Fig.~\ref{fig: Exp uneven terrain results}. In Fig.~\ref{fig: Exp uneven terrain results} (b), the MFT preferable space was well satisfied with jump-free torques. And the robot could track the reference state of the floating-base with acceptable errors.
These results demonstrate the functionalities of MFT-WBC, including trajectory tracking, slippage prevention, jump avoidance of torque command and MFT consideration feature.

In both experiments, the planner and controller had no prior information of push or terrain. The planner never re-planed the trajectories of the floating-base.
Considering the proposed MFT preferable space as the constraint, the MFT-WBC improves the robot's reactive robustness to external impact and perception-less adaptation to the terrain. 
We run our algorithm on the on-board computer (a dual-core Intel Core i7-7600U @ 3.0 GHz). The average computation time of MFT-WBC is around 0.2 ms. However, the maximum computation time can rise to 0.6 ms when the robot state changes significantly.

\section{Conclusions}
\label{sec: CONCLUSIONS}

This paper proposed the MFT-WBC scheme for the robust locomotion of parallel-legged robots. 
The MFT preferable polyhedron was constructed offline and was considered as a soft constraint in the online QP-based WBC to enhance the robot's MFT performance. Appreciate to the offline discretization and polyhedral approximation for the robot MFT indices, the weighted QP-based WBC with 27 optimization variables runs at a 1KHz servo rate with Intel Core i7-7600U (2-core, 3.0 GHz).
We verified the advantage of MFT-WBC in improving the robustness of the system through the comparative simulations. The experimental results showed that, with the proposed MFT-WBC scheme, the parallel-legged robot achieved blindly walking over uneven terrain and push recovery without trajectory re-planning. 

Besides, the offline preprocessing method for MFT formulation proposed in this paper is also applicable to the other robot performance indices in WBC, such as manipulability ellipsoid.
Our future work includes two aspects: 1) extend the proposed method to other performance indices; 2) consider MFT in the locomotion planning.

\section*{ACKNOWLEDGMENT}

This work was supported by UBTECH Robotics and partly supported by the Science and Technology Innovation 2030-Key Project under Grant 2021ZD0201402.



\bibliographystyle{IEEEtran}

\end{document}